\documentclass[sigconf]{acmart}
\usepackage{algorithm}
\usepackage{algpseudocode}
\usepackage{booktabs}
\usepackage{multirow}
\usepackage{multicol}
\usepackage{xspace}
\AtBeginDocument{%
  \providecommand\BibTeX{{%
    \normalfont B\kern-0.5em{\scshape i\kern-0.25em b}\kern-0.8em\TeX}}}

\setcopyright{acmcopyright}
\copyrightyear{2022} 
\acmYear{2022} 
\setcopyright{acmcopyright}\acmConference[WWW '22]{Proceedings of the ACM Web Conference 2022}{April 25--29, 2022}{Virtual Event, Lyon, France}
\acmBooktitle{Proceedings of the ACM Web Conference 2022 (WWW '22), April 25--29, 2022, Virtual Event, Lyon, France}
\acmPrice{15.00}
\acmDOI{10.1145/3485447.3511921}
\acmISBN{978-1-4503-9096-5/22/04}

\newcommand{\mf}[1]{\mathbf{#1}}

\newcommand{\xinput}{{X}_{\mathrm{in}}}
\newcommand{\xinputhat}{\tilde{X}_{\mathrm{in}}}
\newcommand{\xprompt}{{X}_{\mathrm{prompt}}}

\newcommand{\template}{\mathcal{T}}

\newcommand{\labelset}{Y}
\newcommand{\mapping}{\mathcal{M}}

\newcommand{\cls}{\texttt{[CLS]}}
\newcommand{\sep}{\texttt{[SEP]}}
\newcommand{\mask}{\texttt{[MASK]}}

\begin{document}



\title{Ontology-enhanced Prompt-tuning for Few-shot Learning}

\author{Hongbin Ye, Ningyu Zhang}
\authornotemark[1]
\affiliation{%
  \institution{
  Zhejiang University \\ AZFT Joint Lab for Knowledge Engine \\ Hangzhou Innovation Center}
  \city{Hangzhou}
  \country{China}
}
\email{{yehongbin,zhangningyu}@zju.edu.cn}

\author{Shumin Deng}

\affiliation{%
  \institution{
  Zhejiang University \\ AZFT Joint Lab for Knowledge Engine \\ Hangzhou Innovation Center}
  \city{Hangzhou}
  \country{China}
}
\orcid{0000-0002-4049-8478}
\email{231sm@zju.edu.cn}

\author{Xiang Chen}
\affiliation{%
  \institution{
  Zhejiang University \\ AZFT Joint Lab for Knowledge Engine \\ Hangzhou Innovation Center}
  \city{Hangzhou}
  \country{China}
}
\orcid{0000-0002-4049-8478}
\email{xiang\_chen@zju.edu.cn}

\author{Hui Chen, Feiyu Xiong}
\affiliation{%
  \institution{Alibaba Group}
 \city{Hangzhou}
    \country{China}
}
\email{{weidu.ch,feiyu.xfy}@alibaba-inc.com}

\author{Xi Chen}
\affiliation{%
  \institution{Tecent}
 \city{ShenZhen}
    \country{China}
}
\email{jasonxchen@tencent.com}

\author{Huajun Chen}
\authornote{Corresponding author.}
\affiliation{
\institution{
Zhejiang University \& AZFT Joint Lab for Knowledge Engine \& Hangzhou Innovation Center
}
 \city{Hangzhou}
    \country{China}
}
\email{huajunsir@zju.edu.cn}


\begin{abstract}
Few-shot Learning (FSL) is aimed to make predictions based on a limited number of samples. Structured data such as knowledge graphs and ontology libraries has been leveraged to benefit the few-shot setting in various tasks. However, the priors adopted by the existing methods suffer from challenging knowledge missing, knowledge noise, and knowledge heterogeneity, which hinder the performance for few-shot learning. In this study, we explore knowledge injection for FSL with pre-trained language models and propose ontology-enhanced prompt-tuning (OntoPrompt). Specifically, we develop the ontology transformation based on the external knowledge graph to address the knowledge missing issue, which fulfills and converts structure knowledge to text. We further introduce span-sensitive knowledge injection via a visible matrix to select informative knowledge to handle the knowledge noise issue. To bridge the gap between knowledge and text, we propose a collective training algorithm to optimize representations jointly. We evaluate our proposed OntoPrompt in three tasks, including relation extraction, event extraction, and knowledge graph completion, with eight datasets. Experimental results demonstrate that our approach can obtain better few-shot performance than baselines.
\end{abstract}





\maketitle

\section{Introduction}

Recent years have witnessed the success of deep neural networks, however, open issues remain as they are still prone to labeled data in practice and face unignorable challenges owing to the variations of domains, language, and tasks.
These drawbacks lead to the research of an important technique, \textit{few-shot learning} (FSL), which can significantly improve the learning capabilities of machine intelligence and practical adaptive applications by accessing only a small number of labeled examples.
Over the past few years, FSL has been introduced in a wide range of machine learning tasks, such as relation extraction \cite{DBLP:conf/emnlp/ZhangDSCZC18,DBLP:conf/naacl/ZhangDSWCZC19,DBLP:conf/coling/YuZDYZC20,DBLP:journals/corr/abs-2104-07650,DBLP:conf/naacl/LiuFTCZHG21}, event extraction \cite{DBLP:conf/acl/TongXWCHLX20,DBLP:conf/acl/DengZLHTCHC20} and knowledge graph completion \cite{DBLP:conf/aaai/ZhangZWLLSJY21}.

To address the few-shot issue, on the one hand, researchers apply the meta-learning strategy to endow the new model the ability to optimize rapidly with the existing training knowledge or leverage transfer learning to alleviate the challenge of data-hungry \cite{DBLP:conf/www/ZhangDSCZC20,DBLP:journals/corr/abs-2112-10006}. 
Benefiting from the self-supervised pre-training on the large corpus, the pre-train—fine-tune paradigm has become the de facto standard for natural language processing (NLP). It can perform well on downstream tasks with a small amount of task-specific data. 
On the other hand, structured data such as knowledge systems and ontology libraries can be modeled under the few-shot setting. 
Note that those prior knowledge are mostly abstract summaries of human experience and can provide vital support for FSL in various domains \cite{DBLP:conf/www/DengZZCPC19,DBLP:journals/corr/abs-2106-06410,DBLP:journals/corr/abs-2112-01404,zhang2021alicg}. 
In this paper, we focus on injecting knowledge for few-shot learning, and there are still several nontrivial challenges as follows:

\begin{itemize}
\item  \textbf{Knowledge Missing.} Knowledge injection may not be able to retrieve task-relevant facts due to the incompleteness of the external knowledge base and thus provides no useful or even irrelevant information to downstream tasks. 
How to enrich task-relevant knowledge for tasks is an important issue. 

\item  \textbf{Knowledge Noise.} Previous studies \cite{DBLP:journals/corr/abs-2009-13964,DBLP:conf/aaai/LiuZ0WJD020,DBLP:conf/ijcai/ZhangDCCZZC21} have demonstrated that not all knowledge is beneficial for downstream tasks, and an indiscriminate injection of knowledge may lead to negative knowledge infusion, which is detrimental to the performance of downstream tasks. 
Thus, context-sensitive and task-specific knowledge selection is critical for knowledge-enhanced learning.

\item  \textbf{Knowledge Heterogeneity.} The language corpus of downstream tasks is quite different from the injected knowledge leading to two individual vector representations \cite{DBLP:conf/acl/ZhangHLJSL19}.
How to design a special joint training objective to fuse knowledge information is another challenge. 

\end{itemize}

\begin{figure*}[!t]
\centering
\includegraphics[height=5cm]{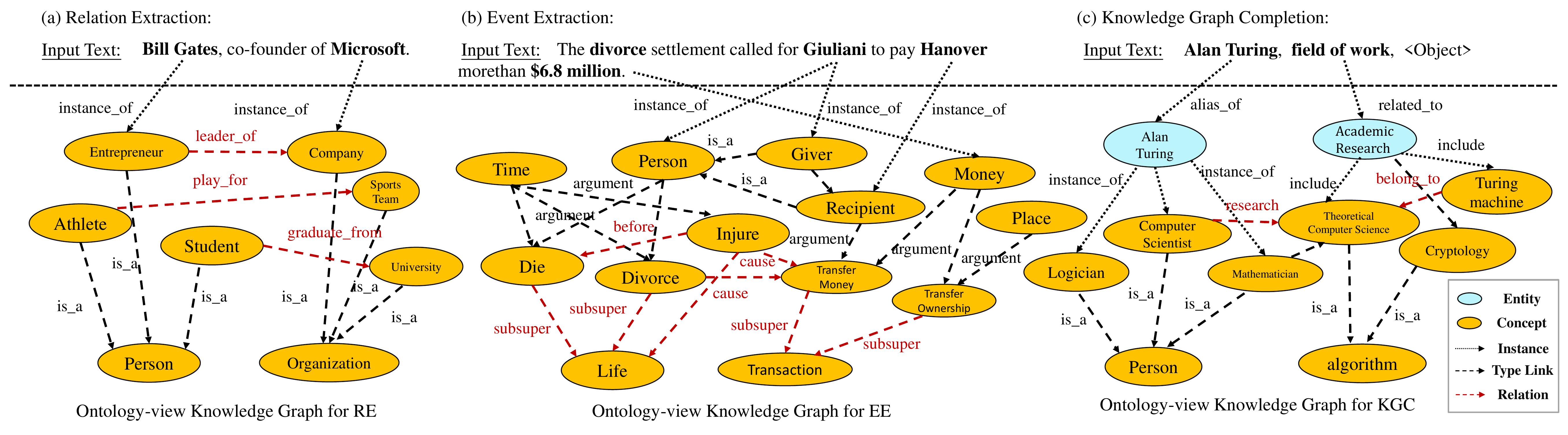}
\caption{Ontology knowledge injection for relation extraction (RE), event extraction (EE) and knowledge graph completion (KGC) (Best viewed in color.).}
\label{fig:intro}
\end{figure*}

In this study, to address the above-mentioned issues, we explore knowledge injection for FSL with pre-trained language models (PLMs) and propose ontology-enhanced prompt-tuning (OntoPrompt). 
Note that pre-trained language models can extract plenty of statistical information from large amounts of data. 
And external knowledge, such as knowledge graphs, is the outcome of human wisdom, which can be good prior to the modeling of statistics. 

\textbf{Firstly}, we propose \emph{ontology transformation} to enrich and convert structure knowledge to text format. 
Specifically, we utilize pre-defined templates to convert knowledge to text as prompts.
Prompt-tuning can reduce the gap between the pre-training model tasks and downstream tasks. 
For example, given a  sentence $\textbf{s}$, "Turing entered King's College, Cambridge in 1931, and then went to Princeton University to study for a doctorate," we can pack them into a knowledgeable prompt based on ontology: "$\textbf{s}$. Turing [MASK] King's College, Cambridge."
PLMs should predict label tokens at the masked position to determine the label of the input.
Note that \emph{ontology as prompt} places knowledge of entities/spans
into the input text, which is model-agnostic and can be plugged into any type of pre-trained language model.

\textbf{Second}, we propose \emph{span-sensitive knowledge injection} to select informative knowledge and mitigate injecting noise. 
Since not all external knowledge is advantageous for the downstream task, and some irrelevant and noisy knowledge may lead to the changes in the meaning of the original sentence, we leverage a visible matrix based on spans and their corresponding external knowledge to guild the knowledge injection. 
In this way, not all tokens in the input sentences will be influenced by external knowledge. 

\textbf{Third}, we propose a \emph{collective training} algorithm to jointly optimize representations. 
Note that the injected external knowledge should be associated with the surrounding context; we add some learnable tokens with random initialization and optimize those tokens as well as injected ontology tokens with language model fixed. 
Inspired by the previous study \cite{DBLP:journals/corr/abs-2109-04332} that prompt-tuning in the low-data regime is unstable and may obtain poor performance, we further optimize all parameters to collective train the ontology text and input text representations.

Finally, we evaluate our OntoPrompt with three tasks: relation extraction, event extraction, and knowledge graph completion. 
We leverage the dataset-related ontologies as external knowledge and conduct experiments on eight datasets in the few-shot setting.
Experimental results illustrate that our proposed approach can obtain better performance. 
It should be noted that our approach is model-agnostic, and therefore orthogonal to existing pre-trained language models. 





\section{Related Work}

\subsection{Knowledge-enhanced Learning}

Pre-training fine-tuning has become a default paradigm for natural language processing. 
However, the performance of the knowledge-driven downstream task (for example, question answering or relation extraction) is dependent on structured relational knowledge; thus, the direct fine-tuning of pre-trained LMs yield suboptimal results. 
Thus, external knowledge graphs have been considered as an indispensable part of language understanding \cite{DBLP:journals/corr/abs-2201-03335}, which has inspired knowledge-aware models such as ERNIE \cite{zhang2019ernie}. 
To integrate the ontology knowledge, \cite{ACL2018_ZSEE} propose to tackle the zero-shot event detection problem by mapping each event mentioned to a specific type in a target event ontology.
\cite{DBLP:conf/acl/DengZLHTCHC20} propose an event detection framework based on ontology embedding with event correlations, which interoperates symbolic rules with popular deep neural networks. 
\cite{DBLP:conf/www/GengC0PYYJC21}  propose a novel ZSL framework called OntoZSL which not only enhances the class semantics with an ontological schema but also employs an ontology-based generative model to synthesize training samples for unseen classes. 
\cite{DBLP:conf/acl/XiangZCCLZ21} propose an ontology-guided entity alignment method named OntoEA, where both knowledge graphs and their ontologies are jointly embedded, and the class hierarchy and the class disjointness are utilized to avoid false mappings.

However, the dilemma of knowledge missing, knowledge noise, and knowledge heterogeneity have not been addressed. 
Concretely, \cite{DBLP:conf/aaai/LiuZ0WJD020}  propose K-BERT, which utilizes soft-position and visible matrix to limit the impact of knowledge. 
\cite{DBLP:journals/corr/abs-2009-13964} propose CokeBERT, which can dynamically select and embed knowledge context according to textual context for PLMs to avoid the effect of redundant and ambiguous knowledge in knowledge graphs that cannot match the input text.
\cite{DBLP:conf/aaai/BianH0021} propose knowledge-to-text transformation to benchmark commonsense question answering. 

Different from their approaches, we integrate ontology knowledge into pre-trained language model fine-tuning with prompts. 
We propose a novel ontology transformation to enrich the missing knowledge and utilize span-sensitive knowledge injection to mitigate the noisy knowledge. 
We further optimize those heterogeneous representations with collective training.

\subsection{Few-shot Learning}

Few-shot learning aims to improve the learning capabilities for machine intelligence, and practical adaptive applications with only a small number of training instances \cite{DBLP:conf/www/ZhangDSCZC20}. 
Our proposed approach corresponds to the other few-shot methods, including: 
(1) Intermediate training \cite{DBLP:journals/corr/abs-1811-01088,DBLP:conf/emnlp/YinRRSX20}, which supplements the pre-trained LMs with further training on the data-rich supervised tasks.
(2) Meta-learning \cite{DBLP:conf/aaai/DengZSCC20,DBLP:conf/wsdm/DengZKZZC20,DBLP:conf/coling/YuZDYZC20}, in which the quantities of the auxiliary tasks are optimized. 
(3) Semi-supervised learning \cite{DBLP:conf/nips/XieDHL020}, which leverages unlabeled samples.

\subsubsection{Relation Extraction} 

Relation extraction aims to identify the relation between entity pairs based on a given contextual text \cite{zhang2021document,10.1007/978-981-16-6471-7_4}. 
In order to reduce the cost of labeling, previous studies utilize distant supervision based on the knowledge graph to generate labeled examples automatically \cite{9537684,ZHANG2022115806}. 
More recent few-shot relation extraction approaches leverage prototype network \cite{DENG2021107584}, multi-level matching and aggregation, relational twin network and meta-learning \cite{DBLP:conf/coling/DongYXGHLLLS20}.

\subsubsection{Event Extraction}

\begin{figure*}[!t]
\centering
\includegraphics[height=10.5cm]{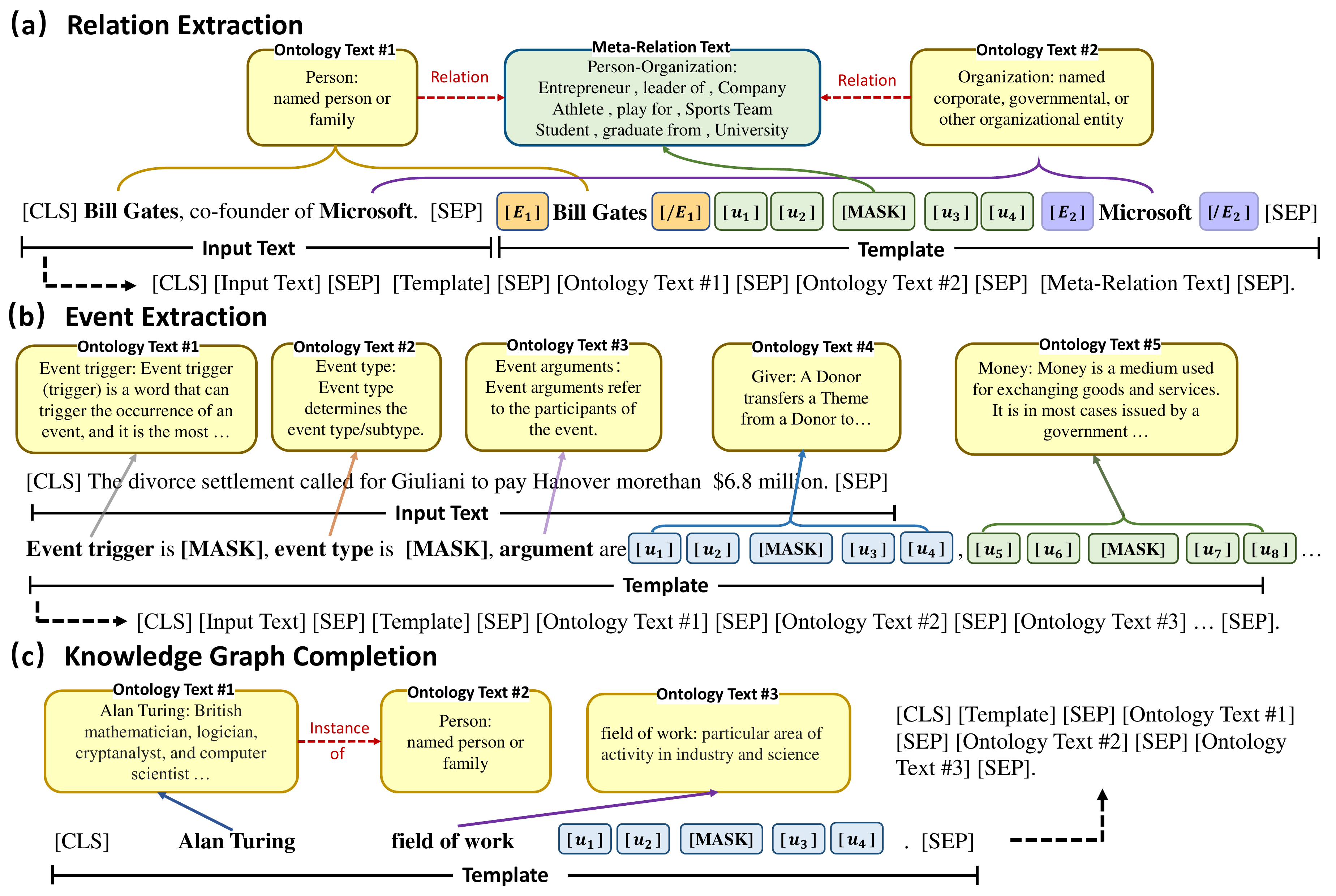}
\caption{Illustration of OntoPrompt for relation extraction (RE), event extraction (EE) and knowledge graph completion (KGC). Those texts in blue and green squares are converted knowledgeable texts from ontology.    
    }
\label{fig:model}
\end{figure*}

Event extraction aims to identify event triggers with arguments from text, which is generally formulated as a classification or structured prediction problem \cite{lou2021mlbinet}. 
To address the few-shot issue for even extraction, \cite{DBLP:conf/acl/LiuCHL016} utilize external knowledge from frame net and propose a global inference approach to improve the event detection performance. 
More recently, \citep{DBLP:conf/emnlp/LiuCLBL20,DBLP:conf/acl-louhi/WangWL20,DBLP:journals/corr/abs-2110-00479} reformulate the event extraction task as machine reading comprehension (MRC), which can achieve better  performance in few-shot setting than vanilla fine-tuning.
Although MRC has the same wide range of application scenarios as our method, we automatically construct templates through ontology knowledge. 

\subsubsection{Knowledge Graph Completion}

Knowledge graph completion can be achieved by link prediction of the knowledge graph \cite{DBLP:conf/www/ZhangPWCZZBC19,DBLP:journals/corr/abs-2201-05575}. 
Previous researchers  have proposed various link prediction methods which  encode entities and relations into the low-dimensional continuous space, such as  TransE \cite{DBLP:conf/nips/BordesUGWY13}, TransR \cite{DBLP:conf/aaai/LinLSLZ15}, TransD \cite{DBLP:conf/acl/JiHXL015}, TransH \cite{DBLP:conf/aaai/WangZFC14}, etc. 
DistMult \cite{DBLP:journals/corr/YangYHGD14a} analyze the matrix and simplify the RESCAL approach, while ComplEx \cite{DBLP:conf/icml/TrouillonWRGB16} extend DistMult to the complex domain.
RotatE \cite{DBLP:conf/iclr/SunDNT19} define each relation as the transformation and rotation from the source entity to the target entity in the complex space. 
KG-BERT \cite{DBLP:journals/corr/abs-1909-03193} take the description of the entities and relations in the triple as input and leverage BERT to calculate the score of the triple. 
For few-shot knowledge graph completion, lots of studies \cite{DBLP:conf/www/ZhangDSCZC20,DBLP:conf/aaai/ZhangYHJLC20,DBLP:conf/emnlp/ShengGCYWLX20} have been proposed.
\cite{DBLP:conf/www/ZhangDSCZC20} propose a general framework called Weighted Relation Adversarial Network (RAN), which utilizes an adversarial procedure to help adapt knowledge/features learned from high resource relations to different but related low resource relations.
\cite{DBLP:conf/aaai/ZhangZWLLSJY21} propose generalized relation learning (GRL), which utilizes semantic correlations between relations to serve as a bridge to connect semantically similar relations.

\subsection{Prompt-tuning}

A new fine-tuning methodology named prompt-tuning has arisen: adapting the pre-trained language model directly as a predictor through completion of a cloze task. 
Prompt-tuning for pre-trained language models is a rapidly emerging field in natural language processing \cite{DBLP:conf/eacl/SchickS21,DBLP:journals/corr/abs-2103-11955,zhang2021differentiable} and have attracted lots of attention. 
Originally  from GPT-3, prompt-tuning has been applied to various of tasks including relation extraction \cite{DBLP:journals/corr/abs-2105-11259}, event extraction \cite{hsu2021event,ye2021learning}, named entity recognition \cite{DBLP:conf/acl/CuiWLYZ21,chen2021lightner}, entity typing \cite{DBLP:journals/corr/abs-2108-10604}, and so on.    
To facilitate the labor-intensive prompt engineering,  \cite{DBLP:conf/emnlp/ShinRLWS20}  propose AUTOPROMPT, which can search prompts based on a gradient method to select label words and templates.
More recent works including P-tuning \cite{DBLP:journals/corr/abs-2103-10385}, Prefix-tuning \cite{DBLP:conf/acl/LiL20} also propose to  leverage continuous templates, which is more effective than discrete prompt search.

Recently, some studies have tried to integrate external knowledge into prompt designing.
\cite{DBLP:journals/corr/abs-2105-11259}  propose an approach called PTR, which leverages logic rules to construct prompts with sub-prompts for many-class text classification. 
\cite{DBLP:journals/corr/abs-2108-02035} propose an approach to incorporate external knowledge graph into the verbalizer with calibration. 
\cite{DBLP:journals/corr/abs-2104-07650} propose a knowledge-aware prompt-tuning approach that injects knowledge into prompt template design and answer construction.
Different from those approaches, we \textbf{regard prompts as a bridge between text and knowledge} and focus on addressing the issues of knowledge missing, knowledge noise, and knowledge heterogeneity.

\section{Methodology}

\subsection{Preliminaries}

Our approach OntoPrompt is a general framework that can be applied to widespread applications as shown in Figure \ref{fig:model}.
We evaluate our approach in three popular tasks, namely, relation extraction (RE), event extraction (EE), and knowledge graph completion (KGC). 
To inject ontology knowledge, we introduce ontology transformation to enrich and convert structure knowledge into raw texts (ontology text). 
We regard those raw texts as auxiliary prompts and append them to the input sequences and prompt templates. 
Note that those ontology texts, including textual descriptions, can provide semantic information about the ontology of the mentioned entities/spans.
We integrate that knowledge during fine-tuning with span-sensitive knowledge injection to avoid external noise. 
We further introduce collective training to optimize prompts as well as language models jointly. 

To facilitate understanding, we first introduce the general framework with prompt-tuning (\S~\ref{vanilla}), ontology transformation (\S~\ref{template}), then introduce span-sensitive knowledge injection (\S~\ref{span}), and finally introduce collective training (\S~\ref{collective}).

\begin{figure*}[!t]
\centering
\includegraphics[height=5.0cm]{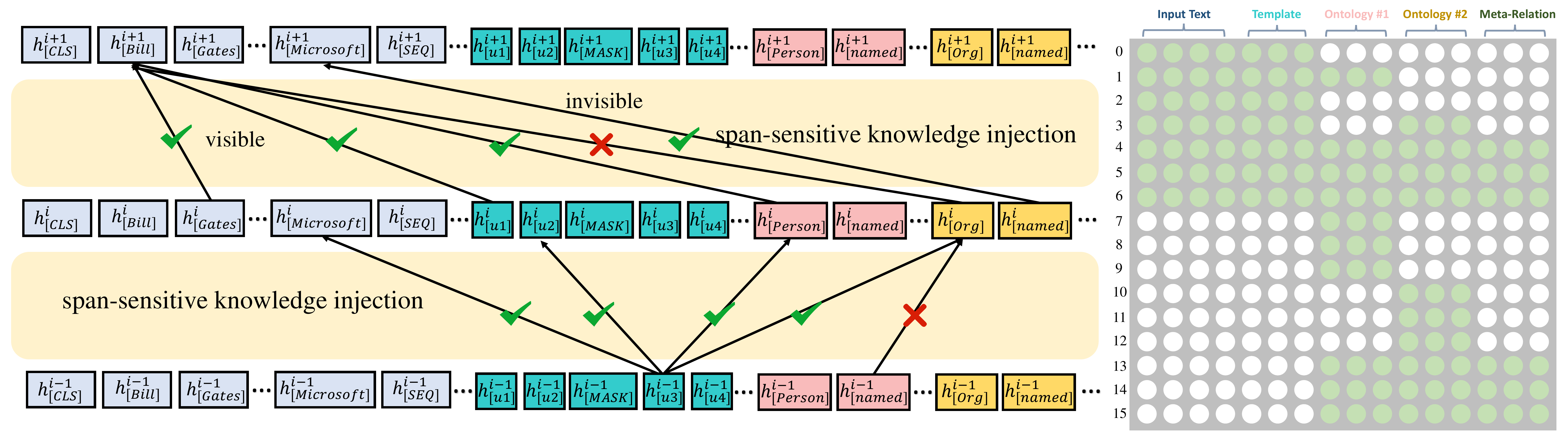}
\caption{Illustration of span-sensitive knowledge injection. 
In this example, Ontology \#1 (red) describes the ontology of the mentioned token starting at position 1, Meta-relation (green) describes the meta-relation path between related ontology pair, and Template (cyan) refers to the template for the prompt. (Best viewed in color.)
    }
\label{fig:att}
\end{figure*}

\subsection{General Framework with Prompt-Tuning}
\label{vanilla}
Let $\xinput=\{x_1, x_2, ..., x_L\}$ be a sentence, where $x_i$ is the $i^{th}$ token in the input sentence and $L$ is the number of tokens. 
Specifically, $\xinput$ is converted to a fixed token sequence $\xinputhat$ and then mapped to a sequence of hidden vectors $\{\mf{h}_k \in \mathbb{R}^d\}$. 
Given the input sequence, $\xinputhat=\cls \xinput \sep$, the vanilla fine-tuning approaches leverage a generic head layer over [CLS] embeddings (e.g., an MLP layer) to predict an output class. 
In this paper, we follow previous prompt-tuning studies \cite{DBLP:conf/acl/GaoFC20} and  use a task-specific pattern string ([Template] $\template$) to coax the model into producing a textual output corresponding to a given class (label token $\mapping(\labelset)$).
Specifically, $\xprompt$ containing one \mask~token is directly tasked with the MLM input as:

\begin{equation}
    \xprompt = \text{\cls $\xinput$ \sep $\template$ \sep}
\end{equation}

When the prompt is fed into the MLM, the model can obtain the probability distribution  $p(\textrm{\tt[MASK]}|(\xprompt)$ of the candidate class, $y \in \labelset$ as:
\begin{equation}
\label{eqn:marginal-prob}
p(y | \xprompt)=\sum_{w \in \mathcal{V}_{y}} p(\textrm{\tt[MASK]} = w | \xprompt)
\end{equation}
where ${w}$ represents the $w^{th}$ label token of class $y$. 

In this paper, we regard ontology text (obtained from external knowledge graph) as auxiliary prompts to inject knowledge and append them to the input sequence templates as shown in Figure \ref{fig:model}. 
We utilize ontology transformation to construct auxiliary prompts which can inject rich task and instance-related knowledge.
To a certain degree, those auxiliary prompts are similar to demonstrations in \cite{DBLP:conf/acl/GaoFC20}; however, auxiliary prompts are not instances in the training set but from external knowledge. 
More details will be introduced in the following sections. 

\subsection{Ontology Transformation}
\label{template}

In this paper, the ontology representation is denoted as
$\mathcal{O}=\{\mathcal{C}, \mathcal{E}, \mathcal{D}\}$, $\mathcal{C}$ is a set of concepts, $\mathcal{E}$ is the connected edge between the ontologies, and $\mathcal{D}$ is the textual description of each ontology (The ontological schema contains a paragraph of textual descriptions, and they are lexically meaningful information of concepts, which can
also be represented by triples using properties, e.g., \emph{rdfs:comment}.). 
The concept set here refers to domain-specific concepts. 
For example, we leverage the type ontology related to the mentioned entity in relation extraction and event extraction. 
We utilize the domain (i.e., head entity types) and range (i.e., tail entity types) constraints in knowledge graph completion. 
 
Regarding the differences in downstream tasks, we leverage different sources of ontology for each task for ontology transformation.  
We first \textbf{extract the ontology of each instance from external knowledge graphs and then transform those ontologies (\emph{rdfs:comment}) into raw texts as auxiliary prompts}. 
 
\subsubsection{Applying to Relation Extraction}

We utilize the MUC (Message Understanding Conference) \cite{vilain1995model} which defines the concept of named entities.
Note that named entities can provide important type information, which is beneficial for relation extraction.
We utilize those definitions as textual descriptions in ontology schema. 
Specifically, we use $[CLS] <Input Text> [SEP] <Template> [SEP] <Ontology Text> [SEP]$ as the final input sequence.
We construct placeholders for entities in the $<Ontology Text>$ and replace those placeholders with external textual descriptions. 
Moreover, we leverage the path between entity pairs from the ontology as meta-relation text to enhance the $<Ontology Text>$ as shown in Figure \ref{fig:model}.
 
We further add learnable tokens as virtual tokens on both sides of the $[MASK]$ so that the model can automatically learn the most suitable words for the prompt. 
Specifically, we use several unused tokens\footnote{The number of tokens is a hyperparameter.}
\textbf{[u1]-[u4]} (e.g., unused or special token in the vocabulary) for virtual tokens. 

\subsubsection{Applying to  Event Extraction}

We follow the work of  \cite{DBLP:conf/acl/DaganJVHCR18} to build the ontology of event extraction. 
We first leverage the ACE event schema \footnote{\url{https://www.ldc.upenn.edu/sites/www.ldc.upenn.edu/files/englishevents-guidelines-v5.4.3.pdf}} as our base event ontology.
The ACE2005 corpus includes the rich event annotations currently available for 33 types. 
However, in real-world scenarios, there may be thousands of types of events. 
In order to facilitate inference in scenarios with limited training samples, we integrate the 33 ACE event types and argument roles with 1,161 frames from FrameNet and construct a bigger event ontology for event extraction. 
We manually map the selected frames to event ontology.
Similarly, we use $[CLS] <Input Text> [SEP] <Template> [SEP] <Ontology Text> [SEP]$ as the default format for the input sequence. 
We construct placeholders for triggers and type in the $<Ontology Text>$. 

Similar to relation extraction, we also leverage the learnable virtual token to enhance the prompt representation. 

\subsubsection{Applying to  Knowledge Graph Completion}

We use the corresponding items obtained from the external Wikidata query as the source of ontology and extract the textual description\footnote{\url{https://www.wikidata.org/}}. 
Following KG-BERT \cite {DBLP:journals/corr/abs-1909-03193}, we regard knowledge graph completion as a triple classification task and concatenate entities and relations as an input sequence.
Similar to the above-mentioned tasks, we use $[CLS] <Input Text> [SEP] <Template> [SEP] <Ontology Text> [SEP]$ as the default input sequence. 

We also use the learnable virtual token to enhance the prompt representation. 
In inference, we rank the output scores according to the maximum probability of sentences predicted by the language model and count the hits of gold standard entities to evaluate the performance of the approach.

\subsection{Span-sensitive Knowledge Injection}
\label{span}

As observed by \cite{DBLP:conf/aaai/LiuZ0WJD020}, excessive knowledge injection may introduce some noise into the original text and cause a performance decay.
In order to solve this problem, we propose span-sensitive knowledge injection as shown in Figure \ref{fig:att}. 
Given input text $X_{in}=[x_1, x_2,..., x_{L}]$ with $L$ tokens, \textbf{we use a visibility matrix to limit the impact of the knowledge injection on the input text}. 
In the language model architecture, the attention mask matrix with self-attention weight is added before the softmax layer. 
Therefore, we modify the attention mask matrix $M$ as follows:

\begin{equation}
M_{ij} =
    \begin{cases}
        0 & x_i, x_j \in x \\
        0 & x_i, x_j \in x^{o} \\
        0 & x_i \in x, x_j \in x^{o} ~\text{and}~ i = p_k \\
        -\infty & \text{otherwise}
    \end{cases}
\end{equation}

where $x_i$ and $x_j$ are tokens from  $x_{in}$ and ontology text, separately. 
$-\infty$  in attention mask matrix $M_{ij}$ blocks token $i$ from attending to token $j$ and $0$ allows token $i$ to attend to token $j$.
$x_i$ can attend to $x_j$ if: both tokens belong to the $x_{input}$, or both tokens belong to the text description of the same ontology $x_o$, or $x_i$ is the token at the span position of entity $e_i$ in $x_{input}$ and $x_j$ is from its ontology description text $x^{o}$. 
$p_k$ indicates the position of the mentioned span (e.g., entities in relation extraction and knowledge graph completion, triggers, or arguments in event extraction) in the input text.

\subsection{Collective Training}
\label{collective}

Note that \textbf{those injected tokens from ontology transformation should be associated with the input sequence}.
Thus, we introduce collective training to optimize ontology tokens and input sequence.
First, we initialize the ontology tokens with real word embeddings and optimize those ontology tokens with the language model fixed. 
Then, we optimize all the parameters of the model, including the language model and ontology tokens. 
Note that our model does not rely on the specific model architecture; thus, it can be plugged into any kind of pre-trained language model like BERT, BART.

\section{Experiments}

In this section, we introduce the extensive experiments of OntoPrompt in three tasks: relation extraction, event extraction, and knowledge graph completion. 
The experimental results show that OntoPrompt can perform better than baselines in both the standard supervised and few-shot settings.

\subsection{Datasets}

\begin{table}[!t]
    \caption{Relation Extraction Dataset statistics.}
        \label{table:1}
    \centering
    \resizebox{0.48\textwidth}{!}{
        \begin{tabular}{cccc}
            \toprule
            \textbf{Dataset}  & \textbf{\#Relations} & \textbf{\#Triples} & \textbf{\#No-relation}  \\
            \midrule
            SemEval-2010 Task 8
                                               & 19            & 10,717              & 17.4\%                  \\
            TACRED-Revisit
                                               & 42            & 106,264              &79.5\%                  \\
            Wiki80
                                               & 80            & 56,000              & -                 \\
            DialogRE
                                              & 36            & 1,788              & -                 \\

            \bottomrule
        \end{tabular}
    }
\end{table}%

\begin{table}[!t]
    \caption{Knowledge Graph Completion Dataset statistics.}
        \label{table:2}
    \centering
    \resizebox{0.48\textwidth}{!}{
        \begin{tabular}{cccccc}
            \toprule
            \textbf{Dataset}  & \textbf{\#Relations} & \textbf{\#Entities} & \textbf{\#Train}  & \textbf{\#Dev} & \textbf{\#Test}\\
            \midrule
            FB15k-237      & 237        & 14,541           & 1,896 
            & 17,535          &2,000
            \\    
            WN18RR      & 18        & 40,943           & 88
            & 3,034         &3,134
            \\    
            UMLS     & 46        & 135          & 329 
            & 652          &661
            \\

            \bottomrule
        \end{tabular}
    }
\end{table}%

\begin{table}[!tb]
    \caption{Results on relation extraction dataset
     for OntoPrompt (F1 score).
     We report the results of fine-tuning BERT\_LARGE with entity markers. We use $K = 8, 16, 32$ (\# examples per class). $Full$ represents the full training set is used. }
     
    \label{tab:RE}
    \small
    \footnotesize
    \centering
    \resizebox{0.48\textwidth}{!}{
    \begin{tabular}{c | c | c | c | c | c}
        \toprule

        \textbf{Dataset} & \textbf{Model} & \textbf{$K = 8$} & \textbf{$K = 16$} & \textbf{$K = 32$} & \textbf{$Full$} \\
                \midrule
        
        \multirow{3}*{SemEval} 
        & Fine-tuning & 24.8 & 43.5 & 63.9 & 87.4 \\
        & GDPNET & 25.3 & 43.5 & 65.2 & 88.0 \\
        & OntoPrompt
        & \textbf{52.6} \tiny{\color{red}{(+27.3)}}  
        & \textbf{65.0} \tiny{\color{red}{(+21.5)}}
        & \textbf{83.0} \tiny{\color{red}{(+17.8)}}
        & \textbf{89.1} \tiny{\color{red}{(+1.1)}} \\

        \midrule
        
        \multirow{3}*{TACRED-\newline
        Revisit} 
        & Fine-tuning & 7.4 & 15.3 & 25.4 & 74.8 \\
        & GDPNET & 7.9 & 17.8 & 26.1 & 77.2 \\
        & OntoPrompt  
        & \textbf{28.8} \tiny{\color{red}{(+20.9)}}  
        & \textbf{33.1} \tiny{\color{red}{(+15.3)}}  
        & \textbf{34.8} \tiny{\color{red}{(+8.7)}}  
        & \textbf{78.2} \tiny{\color{red}{(+1.0)}} \\

        \midrule
    
        \multirow{3}*{WiKi80} 
        & Fine-tuning & 46.1 &60.5 & 70.1 & 85.1 \\
        & GDPNET & 47.4 & 62.3 & 70.5 & 87.0 \\
        & OntoPrompt
        & \textbf{68.7} \tiny{\color{red}{(+21.3)}}  
        & \textbf{75.6} \tiny{\color{red}{(+13.3)}}  
        & \textbf{79.1} \tiny{\color{red}{(+8.6)}} 
        & \textbf{87.9} \tiny{\color{red}{(+0.9)}} \\

        \midrule
        
        \multirow{3}*{DialogRE} 
        & Fine-tuning & 29.3 & 41.1 & 49.5 & 58.3 \\
        & GDPNET & 20.1 & 42.5 & 49.7 & 65.2 \\
        & OntoPrompt
        & \textbf{41.5} \tiny{\color{red}{(+12.2)}}  
        & \textbf{47.3} \tiny{\color{red}{(+4.8)}}
        & \textbf{52.4} \tiny{\color{red}{(+2.7)}}
        & \textbf{66.1} \tiny{\color{red}{(+0.9)}} \\

        \bottomrule
        
    \end{tabular}}
\end{table}

\begin{table*}[!t]
    \caption{\textbf{Knowledge graph completion result (8-shot) on WN18RR, FB15K-237 and UMLS. The best score is in bold.}}
    \label{tab:LP}
    \centering
        \resizebox{0.65\textwidth}{!}{
    \begin{tabular}{@{}c|cc|cc|cc@{}}
    \toprule
    \multirow{2}{*}{Model}               & \multicolumn{2}{c}{WN18RR}           & \multicolumn{2}{c}{FB15K-237 (mini)}        & \multicolumn{2}{c}{UMLS}       \\ \cline{2-7} 
                       & MR   $\downarrow$           & Hit@10 $\uparrow$  & MR   $\downarrow$           & Hit@10 $\uparrow$  & MR   $\downarrow$           & Hit@10 $\uparrow$   \\ \hline
    TransE~\cite{DBLP:conf/nips/BordesUGWY13}   & 19313.0           & 0.0004 & 5847.4   & 0.0925     & 38.5           & 0.3271 \\
    TransR~\cite{DBLP:conf/aaai/LinLSLZ15}    & 20861.0           &0.0023 & 5970.5   & 0.0903     & 43.2            &0.3079  \\
    TransD~\cite{DBLP:conf/acl/JiHXL015}    & 20202.5            & 0.0006 & 5923.1   & 0.0901    & 43.7            & 0.3030 \\
    TransH~\cite{DBLP:conf/aaai/WangZFC14}    & 19272.9           & 0.0015 & 6102.3   & 0.0802     & 41.4            & 0.3166  \\
    DistMult~\cite{DBLP:journals/corr/YangYHGD14a}    & 20671.6       & 0.0003 & 5791.7   & 0.0059    & 59.3            & 0.1716   \\
    Complex~\cite{DBLP:conf/icml/TrouillonWRGB16}    & 20318.6            & 0.0000 & 6451.8   & 0.0046    & 61.3            & 0.2260  \\
    RotatE~\cite{DBLP:conf/iclr/SunDNT19}    & 20162.1            & 0.0003 & 7365.8   & 0.0066    & 68.4            & 0.0526 \\
    KG-BERT~\cite{DBLP:journals/corr/abs-1909-03193}    & 2937.2            & 0.1175 & 2023.4   & 0.0451    & 34.6            & 0.3382  \\
    RAN~\cite{DBLP:conf/www/ZhangDSCZC20}    & 3150.2            & 0.0769 & 3320.5   & 0.0072     & 34.2           & 0.3226 \\ 
    GRL~\cite{DBLP:conf/aaai/ZhangZWLLSJY21}    & 2913.3           & 0.0900& 2913.5   & 0.0300     & 34.5           & 0.3312 \\ \hline
    OntoPrompt    & \textbf{1442.6} \tiny{\color{red}{(-1470.7)}}            & \textbf{0.1344} \tiny{\color{red}{(+0.0169)}} & \textbf{714.5} \tiny{\color{red}{(-1308.9)}}  & \textbf{0.111} \tiny{\color{red}{(+0.0185)}}    & \textbf{27.2} \tiny{\color{red}{(-7.0)}}            & \textbf{0.3448} \tiny{\color{red}{(+0.0066)}}  \\ \hline
    \end{tabular}
    }
\end{table*}

\begin{table}[!t]
    \caption{F1 score (\%) on few-shot learning. 
    The reported F1 score refers to the result of the event argument classification (event extraction) in ACE2005.}
    \label{tab:EE}
    \centering
    \setlength{\belowcaptionskip}{-0.5cm}
    \resizebox{0.48\textwidth}{!}{
        \begin{tabular}{cccccc}
            \toprule
            \textbf{Model}                   & \textbf{1\%} & \textbf{5\%} & \textbf{10\%} & \textbf{20\%} & \textbf{$Full$}\\
            \midrule
dbRNN \cite{DBLP:conf/aaai/ShaQCS18}     & -    & 8.1  & 17.2 & 24.1 & 58.7    \\
JMEE \cite{JMEE}      & -    & 8.9  & 20.3 & 28.4 & \textbf{60.3}    \\
DYGIE++ \cite{DBLP:conf/emnlp/WaddenWLH19} 
& -    & 5.3  & 15.7 & 23.4 & 51.4    \\
MQAEE \citep{DBLP:conf/acl-louhi/WangWL20}     & 5.2 & 27.3 & 32.1 & 38.1 &  53.4   \\
TEXT2EVENT \citep{DBLP:conf/acl/0001LXHTL0LC20} & 3.4 & 19.8 & 25.3 & 36.9 & 49.8   \\
            \midrule
OntoPrompt & \textbf{25.6} \tiny{\color{red}{(+20.4)}} & \textbf{40.1} \tiny{\color{red}{(+12.8)}} & \textbf{47.8} \tiny{\color{red}{(+15.7)}}& \textbf{50.1} \tiny{\color{red}{(+12.0)}}& 55.3 \tiny{\color{red}{(-5.0)}}\\
            \bottomrule
        \end{tabular}
    }
\end{table}%

For relation extraction, we choose a variety of datasets to evaluate OntoPrompt, including sentence-level extraction datasets such as TACRED-Revisit \cite{DBLP:conf/acl/AltGH20a}, SemEval-2010 Task 8, Wiki80 and dialogue-level extraction dataset DialogRE \cite{DBLP:conf/acl/YuSCY20}. 
The detailed statistics of each relation extraction dataset are shown in Table~\ref{table:1}.

For event extraction, we evaluate our OntoPrompt model with the ACE 2005 dataset \cite{DBLP:conf/lrec/DoddingtonMPRSW04}, which defines 33 different event types and 35 semantic roles. 
We use the same data split and pre-processing step following \cite{DBLP:conf/emnlp/WaddenWLH19} in which 529/30/40 newswire documents are used for training/dev/test set. 

For knowledge graph completion, we use several standard knowledge graph completion datasets, including UMLS which has various categories, WN18RR based on WordNet , and FB15K-237 based on Freebase. 
FB15K is a subset of the large-scale knowledge graph Freebase. 
In FB15K-237, the triples that can be reversed are deleted because these triples are more challenging to distinguish.
Compared with FB15K, WN18RR has more entities and fewer types of relations. 
Since the UMLS dataset is a vertical domain dataset, there are fewer entities than the previous two knowledge graphs. 
Detailed statics are shown in Table~\ref{table:2}.

\subsection{Settings}
\label{sec:setting}

The proposed model is implemented using Pytorch. 
Our experiments measure the average performance with a fixed set of seeds, $\mathcal{S}_{seed}$, across five different sampled $\mathcal{D}_{train}$ for each task. 
We utilize a grid search over multiple hyperparameters and select the best result as measured on $\mathcal{D}_{dev}$ for each set $\{\mathcal{D}_{train}^s, \mathcal{D}_{dev}\}, s\in \mathcal{S}_{seed}$.
We employ AdamW as the optimizer. 
We conduct experiments with a BERT\_LARGE model for all experiments.
We use special entity markers uniformly to highlight the entity mentions for relation extraction. 
For the few-shot learning, we follow the settings of \cite{DBLP:conf/acl/GaoFC20}, which is different from the N-way K-shot setting.
We construct prompt templates following \cite{DBLP:journals/corr/abs-2012-15723}.

\subsection{Results}
\label{sec:results}

\begin{figure}[!t]
\centering
\includegraphics[height=5.5cm]{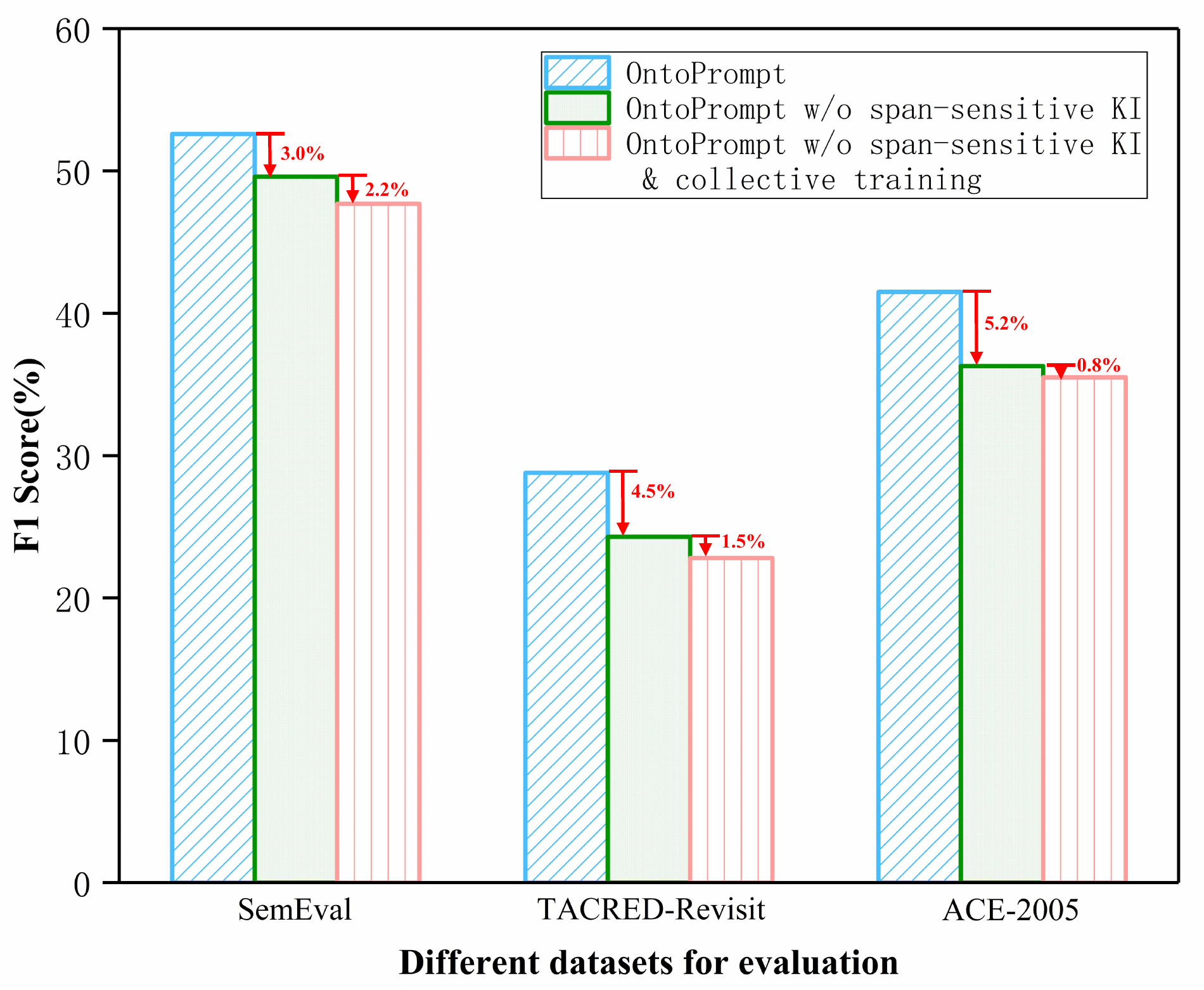}
\caption{Ablation study results of OntoPrompt.}
\label{tab:ablation}
\end{figure}

\begin{table*}[!t]
\caption{
Case study results on the ACE2005 dataset. 
We map the virtual token into the readable natural language (red) in the input examples, demonstrating the intuitiveness of OntoPrompt.
It is worth noting that OntoPrompt can customize a unique virtual token for each prompt based on ontology with different event subtypes. 
In contrast, the virtual tokens of models without knowledge injection are randomly scattered over the entire dataset.
}
\label{tab:study}
\small
\center
\resizebox{1.0\textwidth}{!}{
\begin{tabular}{p{.15\textwidth}|p{.65\textwidth}|p{.1\textwidth}|p{.1\textwidth}}
\toprule
Model                & Input Example                                                                                    & Event Type                                                                & Model F1   \\
\midrule

\textsc{BERT}         & {\texttt{[CLS]}} Everyone, everywhere should have the basic right to elect their government. {\texttt{[SEP]}}                                   & Personnel.Elect & 49.1 \\
\midrule
\textsc{BERT + Ontology}         & {\texttt{[CLS]}} Everyone, everywhere should have the basic right to elect their government. {\texttt{[SEP]}}  Event trigger: Event trigger (trigger) is a word that can trigger the occurrence of an event ...                {\texttt{[SEP]}}                                   & Personnel.Elect & 50.2 \\
\midrule
\textsc{OntoPrompt (-Onto)}         & {\texttt{[CLS]}} Everyone, everywhere should have the basic right to elect their government. {\texttt{[SEP]}}  Event trigger is {\texttt{<Trigger>}}, event type is {\texttt{<Type>}}, argument are \textcolor{red}{[prohibits] [lunged]} {\texttt{[MASK]}} \textcolor{red}{[relies] [lunged]} ...                {\texttt{[SEP]}} Person: named person or family ... {\texttt{[SEP]}}                                         & Personnel.Elect & 52.3 \\
\midrule
\textsc{OntoPrompt (-vir)}         & {\texttt{[CLS]}} Everyone, everywhere should have the basic right to elect their government. {\texttt{[SEP]}}  Event trigger is {\texttt{[MASK]}}, event type is {\texttt{[MASK]}}, argument are {\texttt{[MASK]}} ... {\texttt{[SEP]}}   Person: The Person is conceived of as independent of other specific individuals with whom they have relationships and  ...  {\texttt{[SEP]}}                                               & Personnel.Elect & 52.9 \\
\midrule
\textsc{OntoPrompt (-Both)}         & {\texttt{[CLS]}} Everyone, everywhere should have the basic right to elect their government. {\texttt{[SEP]}}  Event trigger is {\texttt{[MASK]}}, event type is {\texttt{[MASK]}}, argument are {\texttt{[MASK]}} ...        {\texttt{[SEP]}}                                           & Personnel.Elect & 47.5 \\
\midrule

\multirow{3}{*}{\textsc{OntoPrompt}}         & {\texttt{[CLS]}} Everyone, everywhere should have the basic right to elect their government.       {\texttt{[SEP]}}  Event trigger is {\texttt{[MASK]}}, event type is {\texttt{[MASK]}}, argument are \textcolor{red}{[discusses] [receptionist]} {\texttt{[MASK]}} \textcolor{red}{[moaned] [teachings]} ... {\texttt{[SEP]}}  Person: the person elected. The Person is conceived of ...    {\texttt{[SEP]}}                                                                                 & Personnel.Elect  & 
\multirow{3}{*}{\textbf{55.3}}\\
 \cline{2-3}

        & {\texttt{[CLS]}} Here, prisoners were tortured. {\texttt{[SEP]}}   Event trigger is {\texttt{[MASK]}}, event type is {\texttt{[MASK]}}, argument are \textcolor{red}{[focuses] [ethiopia]} {\texttt{[MASK]}} \textcolor{red}{[brownish] [explores]} ... {\texttt{[SEP]}} Place: where the injuring takes
place. Place designates relatively stable bounded ares of the world which have permanent relative ...         {\texttt{[SEP]}}                                                                                & Life.Injure &\\
\cline{2-3}

      & {\texttt{[CLS]}} But it has nothing to do whether we should go to war with Iraq or not. {\texttt{[SEP]}}  Event trigger is {\texttt{[MASK]}}, event type is {\texttt{[MASK]}}, argument are \textcolor{red}{[angrily] [iraqi]} {\texttt{[MASK]}} \textcolor{red}{[hibits] [reddish]} ... {\texttt{[SEP]}} Attacker: The attacking/instigating agent. An assailant physically attacks a Victim  ...         {\texttt{[SEP]}}                                                                                     & Conflict.Attack  &\\

\bottomrule
\end{tabular}
}
\end{table*}

For \textbf{relation extraction}, we compare OntoPrompt with vanilla fine-tuning approach and the most well-performed baseline model GDPNET \cite{DBLP:conf/aaai/XueSZC21} on four datasets. 
As shown in Table~\ref{tab:RE}, it can be found that the OntoPrompt model under the few-shot setting can achieve better performance than the vanilla fine-tuning method in all relation extraction datasets.
In a fully supervised setting, OntoPrompt can obtain an average increase of about 1.0\% compared with GDPNET. 
Note that the implementation of OntoPrompt is relatively simple.
We believe that our model can be plugged into different models to promote the performance of relation extraction tasks in more complex scenarios.

For \textbf{event extraction}\footnote{We only report the performance of event argument classification due to page limit.},  from Table \ref{tab:EE}, we report the F1 score results of event extraction in extremely low-resource scenarios (training with less than 20\% data, with the similar setting to \cite{DBLP:conf/emnlp/LiuCLBL20}. 
Notably, OntoPrompt yields advantages in few-shot event extraction. 
To be specific, OntoPrompt can obtain 25.6\% F1 with 1\% data, in comparison to 5.2\% in MQAEE and 3.4\% in TEXT2EVENT. 
Although the performance of OntoPrompt on the full sample is slightly weaker than that of JMEE, which relies on external data augmentation, we believe that OntoPrompt can effectively identify triggers and arguments with less data dependence.
Besides, we notice that OntoPrompt has a faster convergence speed (See details in appendix).
Compared with MQAEE and TEXT2EVENT, which need a certain amount of data to fine-tune the language model, OntoPrompt only leverages 20\% of the training data to achieve similar satisfactory performance.

For \textbf{knowledge graph completion}\footnote{Due to the huge computation in ranking during the inference stage, we construct and evaluate the models on a  small test set called FB15K-237 (mini), baselines are implemented by \url{https://github.com/thunlp/OpenKE}}, from Table~\ref{tab:LP}, we report the performance of multiple knowledge graph completion approaches in the few-shot setting (8 samples per relationship, 8-shot). 
The experimental results show that OntoPrompt can achieve the best performance, proving our proposed model's superiority. 
Although OntoPrompt has a slight improvement in the few-shot setting on UMLS, it has an increase of 6.6\% with the hits@10  on FB15K-237 compared to KG-BERT. 
We think the improvement of FB15K-237 is mainly due to OntoPrompt's ability to fully utilize the implicit fact knowledge obtained from external ontology. 

Although we evaluate OntoPrompt with the above three tasks, our proposed approach can also be applied to other tasks such as text classification, question answering with suitable ontologies.

\subsection{Ablation Study}
To further prove the effects of different modules in OntoPrompt, we conduct the ablation study and report the experimental results in  Figure~\ref{tab:ablation}. 
\emph{w/o span-sensitive KI} indicate the model without the visible matrix; thus, all tokens can see each other. 
\emph{w/o span-sensitive KI \& collective training} refers to the model without the visible matrix, and all parameters are optimized. 
We observe that all models without each component have a performance decay, which demonstrates the effectiveness of our approach.

\subsection{Case Study}

To further analyze the collective training of virtual tokens,  we conduct the nearest neighbor word embedding search for the virtual token in the sentence to project the best optimized virtual token into a readable natural language.
From Table \ref{tab:study}, we can observe the following findings:

1) The OntoPrompt can customize a unique virtual token for each ontology in different event subtypes, while the virtual tokens of the model without ontology knowledge are random tokens scattered in the entire dataset. 
This indicates that injecting external knowledge can benefit prompt template representation.

2) The integration of ontology knowledge can improve the base pre-trained model to varying degrees, e.g., BERT (+1.1\%) and OntoPrompt (+3.0\%), which further prove that the ontology knowledge is helpful to the downstream tasks such as event extraction. 

3) The removal of virtual template tokens reduces our OntoPrompt performance (-2.4\%). 
We believe that the virtual token can learn the implicit information of the task-specific knowledge and adjust the appropriate prompt according to the context. 
At the same time, the removal of the virtual template token and the ontology knowledge module both will result in a more severe performance decay (-7.8\%), which demonstrates that the virtual template token and the ontology knowledge information both have a positive impact on the model performance.

\subsection{Discussion}
Notably, our  OntoPrompt can be viewed as an approach to leverage prompts as a bridge to inject knowledge. 
Previous studies such as RAG \cite{DBLP:conf/nips/LewisPPPKGKLYR020} has the same intuition to retrieve and concatenate relevant texts as knowledge. 
That knowledge can be directly revised and expanded, and accessed knowledge can be inspected and interpreted.
Apart from those approaches, we argue that prompts transformed from ontology can consist of more density knowledge.
Moreover, we utilize span-sensitive knowledge injection to filter noisy information. 
However, our model cannot handle those \textbf{complex} or \textbf{structure} knowledge such as OWL reasoning rules, description logic and represent them as raw texts. 
We will leave this for future works.

\section{Conclusion and Future Work}

We focus on ontology knowledge injection and propose an ontology-enhanced prompt-tuning (OntoPrompt) approach, which can be applied to relation extraction, event extraction, and knowledge graph completion tasks.
Experimental results illustrate that OntoPrompt can obtain better results than baselines in eight datasets. 
The method in this paper verifies the effectiveness of ontology knowledge as prompt guidance.

In the future, we plan to apply our approach to more applications such as text generation, question answering. 
We will also try to combine the proposed method with semi-supervised learning algorithms to make better use of large amounts of unlabeled data.
In addition, we will try to inject the more complex knowledge such as symbolic rules in the knowledge graph into the proposed model to construct more powerful prompt templates.

\section*{Acknowledgments}
We  want to express gratitude to the anonymous reviewers for their hard work and kind comments. 
This work is funded by  National Key R\&D Program of China (Funding No.SQ2018YFC000004), NSFC91846204/NSFCU19B2027, Zhejiang Provincial Natural Science Foundation of China (No. LGG22F030011), Ningbo Natural Science Foundation (2021J190), and Yongjiang Talent Introduction Programme.

\bibliographystyle{ACM-Reference-Format}
\bibliography{sample-base}

\appendix

\section{Convergence Analysis}

\begin{figure}[!h]
\centering
\includegraphics[height=4.8cm]{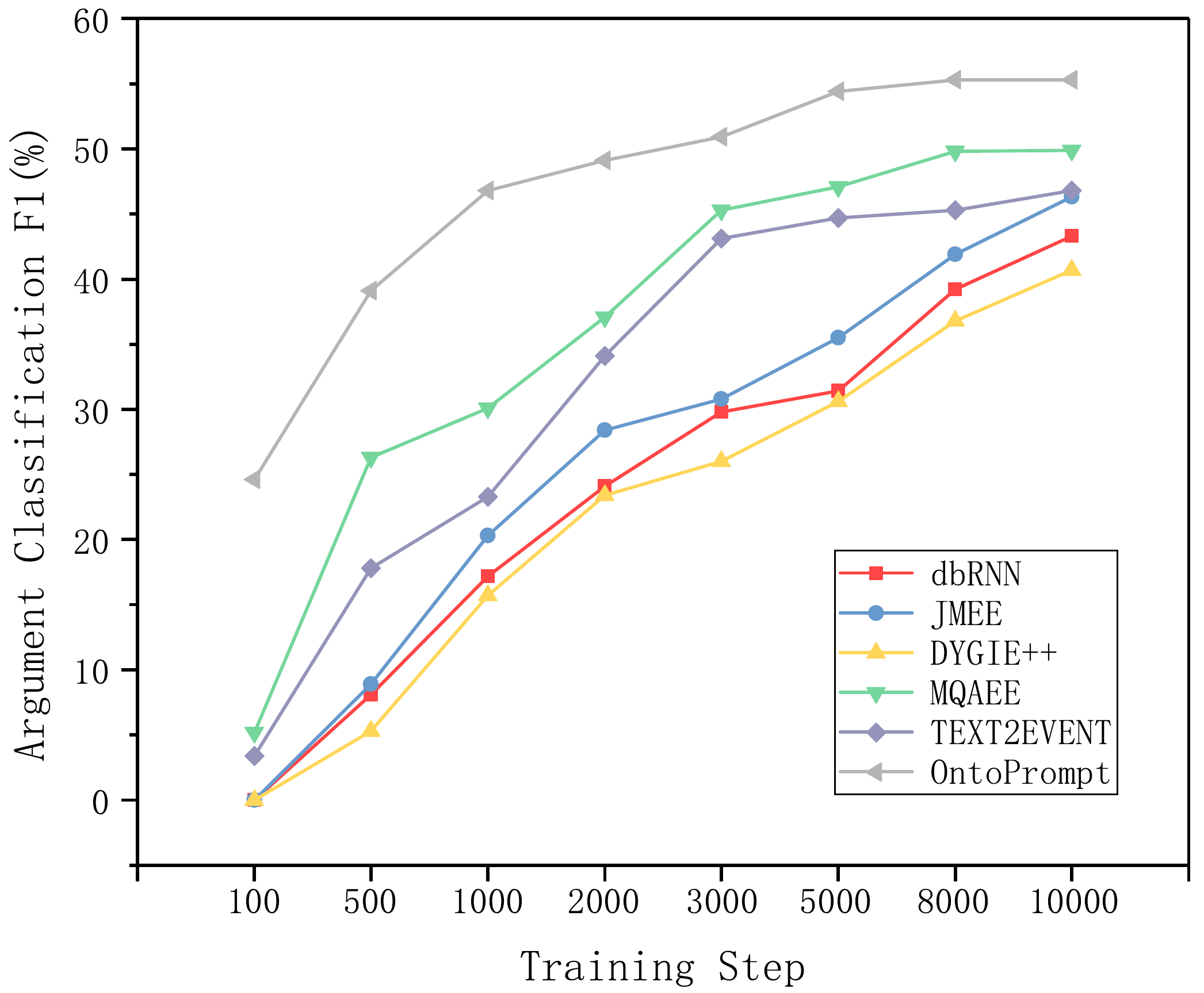}
\caption{Convergence Analysis Results of OntoPrompt.}
\label{tab:conver}
\end{figure}


\end{document}